\def\year{2020}\relax
\documentclass[letterpaper]{article} 
\usepackage{aaai20}  
\usepackage{times}  
\usepackage{helvet} 
\usepackage{courier}  
\usepackage[hyphens]{url}  
\usepackage{graphicx} 
\usepackage{graphicx}  
\usepackage{tikz}
\usepackage{colortbl}
\usepackage{subcaption}
\usepackage{booktabs}
\usepackage{amsmath}

\newcommand{\citet}[1]{\citeauthor{#1} \shortcite{#1}}
\title{Weakly Supervised POS Taggers\\Perform Poorly on {\em Truly} Low-Resource Languages}

\author{Katharina Kann\thanks{The first two authors contributed equally.}\\
Center for Data Science \\ New York University, USA \\ {\tt kann@nyu.edu}
\And
Oph\'{e}lie Lacroix$^*$\\
Siteimprove, Denmark \\ {\tt ola@siteimprove.com}
\And 
Anders S{\o}gaard \\
Department of Computer Science\\ University of Copenhagen, Denmark\\ {\tt soegaard@di.ku.dk}
}

\begin{document}

\maketitle

\begin{abstract}
Part-of-speech (POS) taggers for low-resource languages which are exclusively based on various forms of weak supervision -- e.g., cross-lingual transfer, type-level supervision, or a combination thereof -- 
have been reported to perform almost as well as supervised ones. However, weakly supervised POS taggers are commonly only evaluated on languages that are very different from \emph{truly} low-resource languages, and the taggers use sources of information, like high-coverage and almost error-free dictionaries, which are likely not available for resource-poor languages.
We train and evaluate state-of-the-art weakly supervised POS taggers for a typologically diverse set of 15 truly low-resource languages. 
 On these languages, given a realistic amount of resources, even our best model gets only less than half of the words right. Our results highlight the need for new and different approaches to POS tagging for truly low-resource languages.
\end{abstract}

\section{Introduction}
Part-of-speech (POS) tagging can be very helpful in many natural language processing (NLP) applications, including machine translation, question answering, or relation extraction, especially for low-resource languages \cite{Sennrich:Haddow:16,Nadejde:ea:17,Currey:Heafield:19}. 
POS taggers assign a syntactic category, i.e., a part of speech, to each token in a sentence, thus providing a rudimentary syntactic analysis of the sentence.
While ambiguity and unseen words make POS tagging non-trivial, supervised POS tagging is often considered a relatively simple problem,
reaching a performance close to inter-annotator agreement \cite[$97.96\%$ accuracy for English]{bohnet-EtAl:2018:Long}. The fact that weakly supervised POS taggers sometimes obtain almost-as-good scores makes POS tagging come across as a task where there is little room for improvement. We show that on the contrary, there is {\em a lot}~of room for improvement.  

The accuracy of POS taggers naturally depends on the quality of available training data, as well as the granularity of the POS inventory, i.e., the size of the tag set. Nevertheless, for almost all resource-rich languages considered in the literature, supervised POS taggers have reported tagging accuracies in the high 90s.\footnote{The CoNLL 2018 shared task \cite{conllsharedtask2018} on universal dependencies reported macro-average F$_1$ scores of up to $96.23\%$ on a set of 61 treebanks in high-resource languages.} 
However,
supervised POS taggers typically require corpora that have been manually created by professional linguistic annotators.\footnote{See \citet{Hovy:ea:15} for experiments with crowdsourcing POS annotation, indicating that sometimes, lay annotators can be used at a small cost in accuracy.} 
In the absence of such corpora, we have to resort to alternative sources of (weak) supervision. One very popular source of such supervision is tag dictionaries \cite{li2012wiki,garrette2013learning,tackstrom2013token,wisniewski2014cross,moore2015,plank-agic-2018-distant}.  

In this paper, we revisit and critically examine existing methods for weakly supervised POS tagging. 
While such approaches have been created to meet the needs of low-resource languages, many experiments with weakly supervised POS taggers have been limited to languages for which
good resources actually exist. 
This has practical reasons: it is easier to obtain translations, dictionaries, and benchmark data for widely studied languages. However, since the resource-rich languages are {\em not}~a representative sample of the world's languages, this means we have no guarantee that results scale to truly low-resource languages.

\textbf{Contributions} We study the performance of 
weakly supervised POS taggers for actual low-resource languages. 
To this end, we consider a language \textit{truly low-resource} for a task when no or almost no resources, i.e., annotated corpora or manually created dictionaries, are available for that task in that language.
Assuming a setting in which we have access to no annotated data in the target language at all,
we investigate several state-of-the art baselines, as well as two cross-lingual transfer variants of a state-of-the-art model for the high-resource case \cite{Plank16multi}, extended to a multi-task learning (MTL) architecture in order to provide additional character-level supervision \cite{kann-etal-2018-character}. 
Our results show that POS tagging is {\em still a difficult problem in the truly low-resource setting}:
{\bf Our strongest baseline} obtains a macro-average accuracy of $\mathbf{39.11\%}$ over 15 languages, even though it was developed specifically for low-resource languages.
{\bf Our best architecture}, based on cross-lingual transfer and character-level MTL, is slightly better, with a macro-average accuracy of $\mathbf{42.25\%}$, but our experiments emphasize the need for additional research on POS tagging for truly low-resource languages.
We provide our preprocessed data for all languages  
under 
\textit{https://bitbucket.org/olacroix/truly-low-resource} 
in order to facilitate future research.

\textbf{Disclaimer} In this study, we do not consider the case where small amounts of labeled data is (also) available for the target language. While it is easy to think that annotating a few hundred sentences should be cheap, developing guidelines, as well as  finding and training annotators for truly low-resource languages is challenging and often impossible. Moreover, we do not look at the potential gains from using contextualized word embeddings from pretrained language models \cite{peters2018deep,devlin2018bert}, since even high-quality language models can be hard to train for truly low-resource languages, because data is scarce and exhibits significant spelling variation, and because language detection software has limited precision for truly low-resource languages \cite{Baldwin:Lui:10}. 

\section{Unsupervised and Weakly Supervised POS Tagging}
\label{sec:related_work}
Since the resources required to train supervised taggers are expensive to create and unlikely to exist for the majority of the world's languages, many unsupervised and weakly supervised methods---the latter often building on the former---have been developed. We now briefly survey this work.

\textbf{Unsupervised POS tagging}
Hidden Markov models (HMMs) and their variants have been standard models for POS induction  
\cite{Merialdo:1994:TET:972525.972526,goldwater-griffiths:2007:ACLMain}. 
\citet{christodoulopoulos2010two} evaluated seven different POS induction systems, from the well-known clustering method of \citet{brown1992class} to \citet{bergkirkpatrick-EtAl:2010:NAACLHLT}, who incorporated linguistically motivated features into HMMs.
Subsequently, \citet{Christodoulopoulos2011Bayesian} proposed a Bayesian multinomial mixture model instead of a sequence model to infer clusters from unlabeled text.
More recent studies include \cite{stratos2016unsupervised} who proposes to use anchor HMMs.

POS tags could then be assigned to clusters, e.g., by using monolingual dictionaries, however, in most studies gold annotated data are used to find the best match of POS-tags, hence reporting upper-bound performance of their systems.
While all approaches mentioned above were reported to perform reasonably, none of
the results were obtained for actual low-resource languages (and, in a low-resource setting such as when using low-resource dictionaries), e.g., $75.5\%$ for English \cite{bergkirkpatrick-EtAl:2010:NAACLHLT},
$74.4\%$ for German, and $61.5\%$ for Arabic \cite{Christodoulopoulos2011Bayesian}.

\textbf{Weakly supervised POS tagging}
Weakly supervised methods for POS tagging tend to rely on the projection of information across word alignments between parallel corpora \cite{
tackstrom2013token,agic2015bible,plank-agic-2018-distant}, cross-lingual features obtained from parallel data, comparable data, or seed bilingual dictionaries \cite{gouws2015simple,fang2017model}, or monolingual tag dictionaries that constrain the search space of unsupervised 
POS induction \cite{li2012wiki,wisniewski2014cross,garrette2013learning}.
Strategies using dictionaries extracted from \textsc{Wiktionary} have been popular for learning from noisily labeled sentences \cite{li2012wiki,wisniewski2014cross}. 
\citet{li2012wiki} proposed a feature-based maximum entropy emission model for learning POS taggers from texts which have been labeled using monolingual dictionaries.
However, they evaluate the performance of their system exclusively on resource-rich 
languages for which such dictionaries cover a large amount of their vocabulary, resulting in performances over $80\%$.
\citet{garrette2013learning}
 proposed a maximum entropy Markov model for learning POS taggers from noisily labeled sentences. They labeled a raw corpus with the help of a manually annotated dictionary that was expanded by label propagation. Notably, the authors did evaluate on actual low-resource languages.
However, to actually achieve reasonable results, they bootstrapped their model with a small set of manually annotated sentences.

Our work is also related to \citet{fang2017model}, in that we {do not} rely on parallel text, since we are interested in a setting in which large amounts of parallel corpora are not available, as might be the case for true low-resource languages. However, in contrast, we do not  assume a small amount of training data for the target language.
Further, our approach is similar to \citet{tackstrom2013token} in creating silver-standard training data from type-level dictionaries, but we make use of the state-of-the-art architecture by \citet{Plank16multi} instead of employing HMMs.
Our study is also  similar to \citet{agic2015bible} in evaluating approaches to weakly supervised POS tagging across many low-resource languages, but differs in that we are interested in methods without need for parallel corpora. 

\section{Model Architecture}
\label{sec:model}
Hierarchical POS tagging long short-term memory networks (LSTMs),
such as the architecture proposed by \citet{Plank16multi}, 
receive both word-level and subword-level input.
They 
perform well, even on unseen words,
due to their ability to associate subword-level patterns with POS tags. 
This 
is important 
in a low-resource setting.
However, hierarchical LSTMs are also very expressive, and thus prone to overfitting.
In order to overcome this, we additionally train our models on 
subword-level auxiliary tasks \cite{kann-etal-2018-character} to regularize the character-level encoding in hierarchical LSTMs.
Such a model is still able to make predictions about unknown words,
but the subword-level auxiliary task should prevent it from overfitting. 

\subsection{Hierarchical LSTMs with Character-Level Decoding} 
For the hierarchical sequence labeling LSTM, we follow \citet{Plank16multi}: 
our subword-level LSTM is bidirectional and operates on the character level \cite{
conf/emnlp/BallesterosDS15}.
Its input is the character sequence of each input word, represented by the embedding sequence 
$c_1, c_2, \dots, c_m$. The final character-based representation of each word is the concatenation of the 
two last LSTM hidden states:
\begin{align}
v_{c, i} = conc(&\text{LSTM}_{c, f}(c_{1:m}), \text{LSTM}_{c, b}(c_{m:1}))
\end{align}
Here, $\text{LSTM}_{c, f}$ and $\text{LSTM}_{c, b}$ denote a forward and backward LSTM, respectively.

Second, a context bi-LSTM operates on the word level.
Like \citet{Plank16multi}, we use the term ``context bi-LSTM'' to denote a bidirectional LSTM which, in order to generate a representation for input element $i$,
encodes all elements up to 
position $i$ with a forward LSTM and all elements from $n$ to $i$ using a backward LSTM.
For each sentence represented by embeddings $w_1, w_2, \dots, w_n$, 
its input are the concatenation of the word embeddings with the outputs of the subword-level LSTM:
$conc(w_1,v_{c, 1}), conc(w_2, v_{c, 2}) \dots, conc(w_n, v_{c, n})$. 
The final representation which gets forwarded to the next part of the network is again the concatenation of the last 
two hidden LSTM states:
\begin{align}
v_{w, i} = conc(&\text{LSTM}_{w, f}(conc(w, v_{c})_{1:i}),\\\nonumber &\text{LSTM}_{w, b}(conc(w, v_{c})_{n:i}))
\end{align}
This is then passed on to a classification layer.

\subsection{Character-Based Seq2Seq Model}
\noindent The second component of our model is based on 
a character-level sequence-to-sequence (seq2seq) architecture. It consists of a bi-LSTM encoder which is connected to an LSTM
decoder \cite{sutskever2014sequence}. 

\textbf{Encoding} The encoder is the character-level bi-LSTM described above, and, thus, yields the representation
\begin{align}
v_{c, i} = conc(&\text{LSTM}_{c, f}(c_{1:m}), \text{LSTM}_{c, b}(c_{m:1}))
\end{align}
for an input word embedded as $c_1, c_2, \dots, c_m$.
Parameters of the character-level LSTM are shared between the sequence labeling and the seq2seq part of our model.

\textbf{Decoding}
The decoder receives the concatenation of the last hidden states, $v_{c, i}$, as input.
In particular, we do not use an attention mechanism \cite{bahdanau2014neural}, since
our goal is not to improve performance on the auxiliary task, but instead 
to encourage the encoder to learn better word representations.
The decoder is trained to predict each output character $y_{t}$ dependent
on $v_{c, i}$ and  previous predictions {$y_1$, ..., $y_{t-1}$} as
\begin{align}
  &p(y_t | \{y_1, ..., y_{t-1}\}, v_{c, i})  = g(y_{t-1}, s_t, v_{c, i})
\end{align}
for a non-linear function $g$ and the LSTM hidden state $s_t$. 
The final softmax output layer is calculated over the character vocabulary of the language.

\subsection{Multi-Task Learning}
The character-level LSTM is  
shared between the sequence labeling and
the seq2seq components of our network. All model parameters, including all embeddings, 
are updated during training.

We want to train our neural model jointly on (i) a low-resource main task, i.e., POS tagging, and (ii) a
character-level auxiliary task, namely word autoencoding. Therefore, we want to maximize the following joint log-likelihood:
\begin{align}
  {\cal L}({\boldsymbol \theta}) ~ =  &\!\!\sum_{(l, s) \in {\cal D}_{\textrm{POS}}} \!\!\log p_\theta\left(l \mid s \right) ~ + \\\nonumber
  &\sum_{(in, out) \in {\cal D}_{\textrm{aux}}} \!\!\log p_\theta\left(out \mid in \right)
\end{align}
Here, ${\cal D}_{\textrm{POS}}$ denotes the POS tagging training data, with $s$ being the input sentence and $l$ the corresponding
label sequence. ${\cal D}_{\textrm{aux}}$ is our auxiliary task training data with examples
consisting of input $in$ and output $out$. 
The set of model parameters $\boldsymbol \theta$ is the union of the set of parameters of the sequence labeling
and the seq2seq part. Parameters of the character-level LSTM are shared between tasks.

\textbf{Word autoencoding (AE)}
Our auxiliary task consists of reproducing a
given input character sequence in the output. 
Thus,  training examples are of the form $w \mapsto w$, where $w \in {\cal V}_L$ for the vocabulary ${\cal V}_L$ of $L$. Word autoencoding has been used as an auxiliary task before, e.g., by \citet{Vosoughi:ea:16}.

\begin{table*}[ht]
\centering
{\small
{\setlength{\tabcolsep}{9pt}
\begin{tabular}{llrrrrrrr}
\toprule
\multicolumn{2}{c}{\textbf{Language}} & \multicolumn{2}{c}{\textbf{Treebank Data} (test)} & \textbf{\textsc{Unimorph}} & \textbf{\textsc{Wikidata}+\textsc{Panlex}} & \multicolumn{2}{c}{\textbf{\textsc{Wikipedia}} (\# tagged)} & \textbf{Embeddings} \\
\cmidrule(lr){1-2} \cmidrule(lr){3-4} \cmidrule(lr){5-5} \cmidrule(lr){6-6} \cmidrule(lr){7-8} \cmidrule(lr){9-9}
code & family & sentences & tokens & \multicolumn{1}{c}{entries} & translations & sentences & tokens & \multicolumn{1}{c}{entries} \\
\midrule
am & AA & 1,095 & 10k & - & 2.7k & 777 & 17.9k & 10k \\
be & IE & 68 & 1.3k & - & 35.3k & 7,385 & 101.9k & 93k \\
br & IE & 888 & 10.3k & - & 12.2k & 9,083 & 112.9k & 39k \\
fo & IE & 1,208 & 10.0k & 45.4k & 2.9k & 9,958 & 144.6k & 12k \\
hsb & IE & 623 & 10.7k & - & 4.6k & 1,858 & 30.2k & 10k \\
hy & IE & 514 & 11.4k & 338k & 65.1k & 3,560 & 71.4k & 47k \\
kmr & IE & 734 & 10.1k & - & 4.6k & 3,225 & 48.3k & 24k \\
lt & IE & 55 & 1.0k & 34.1k & 38.9k & 11,464 & 117.2k & 100k \\
mr & IE & 47 & 0.4k & - & 23.4k & 4,886 & 55.2k & 47k \\
mt & AA & 100 & 2.3k & - & 2.1k & 2,361 & 43.9k & 16k \\
bxr & Mo & 908 & 10.0k & - & 2.7k & 2,308 & 37.8k & 28k \\
kk & Tu & 1,047 & 10.1k & - & 63.5k & 12,273 & 122.4k & 100k \\
ta & Dr & 120 & 2.2k & - & 27.1k & 5,772 & 76.2k & 100k \\
te & Dr & 146 & 0.7k & - & 28.0k & 7,872 & 90.9k & 100k \\
tl & Au & 55 & 0.2k & - & 6.8k & 5,871 & 97.6k & 41k \\
\midrule
de & IE & 1,000 & 21.3k & 179.3k & 90.2k & 12,162 & 195.1k & 100k \\
es & IE & 1,000 & 23.3k & 382.9k & 59.7k & 15,209 & 276.6k & 100k \\
it & IE & 1,000 & 23.7k & 509.5k & 59.7k & 10,254 & 170.0k & 100k \\
pt & IE & 1,000 & 23.4k & 303.9k & 47.9k & 12,674 & 195.2k & 100k \\
sv & IE & 1,000 & 19.1k & 78.4k & 58.8k & 10,243 & 134.5k & 100k \\
\bottomrule
\end{tabular}
}
}
\caption{Resources for our low-resource languages (up) and high-resource languages (down). Language families: Afro-Asiatic (AA), Austronesian (Au), Dravidian (Dr), Indo-European (IE), Mongolic (Mo), and Turkic (Tu).  
}\label{tab-all-stats}
\end{table*}
\section{Weak and Cross-Lingual Supervision}
The model described 
above relies on full supervision.
However, in our setting, we need to rely on alternative resources, including raw corpora and linguistic resources, and cross-lingual transfer to create training examples. 
The raw corpora which we leverage for our approaches consist of cleaned \textsc{Wikipedia} texts for our languages. Preprocessing involves (a) segmentation of text into sentences; (b) tokenization; (c) removal of sentences that include at least one of (i) mostly foreign characters; (ii) mostly symbols or punctuation; (iii) no words from our dictionaries. 
Most of the texts that we extract for our low-resource languages 
are predominantly made up of ambiguous sentences (i.e., most words in these sentences are mapped to more than one POS tag in our dictionaries)
which makes it impossible to extract, for training, only sentences that would be fully (and unambiguously) tagged with our dictionaries.
Numbers of sentences and tokens extracted for each language are shown in Table~\ref{tab-all-stats}.

In order to obtain silver-standard training data for POS tagging in our low-resource languages, we rely on cross-lingual transfer. In particular, we assume the following to be available for each low-resource language: (i) raw text, e.g., the \textsc{Wikipedia} corpora we just described; (ii) a bilingual dictionary $B$, containing translations which consist of pairs of words $(w_l, w_h)$ in the low-resource language $L_l$ and a high-resource language $L_h$, respectively; (iii) large amounts of gold POS-annotated data $D$ in the high-resource language $L_h$; and, optionally, (iv) a monolingual tag dictionary $M$. 
Given those resources, we propose two cross-lingual transfer methods, which we will outlay in the following.

\textbf{Frequency-based annotation (\textbf{FREQ})}
The high-level idea of our first approach is to tag each token with the POS tag which has been most frequently assigned to its high-resource language translations:
\begin{align}
\text{TAG}(w_l) = ~ &t \in \cup_{w_h\in \text{Tr}(w_l)} \text{POS}(w_h) : \nonumber\\
& \text{freq}_{\text{Tr}(w_l)}(t) \geq \text{freq}_{\text{Tr}(w_l)}(s) \nonumber\\
& \forall s \in \cup_{w_h\in \text{Tr}(w_l)}\text{POS}(w_h)
\end{align}
Here, $\text{Tr}(w_l)$ denotes all possible translations of $w_l$ in $B$,  
$\text{POS}(w_h)$ is the set of all attested POS tags of $w_h$ in $D$ and 
$\text{freq}_{\text{Tr}(w_l)}(x)$ is the number of times tag $x$ has been assigned to any word in 
$\text{Tr}(w_l)$ in $D$.
 $B$ is a type-level resource, while the training data $D$ in the high-resource language is token-based, i.e., words are tagged in context.

A word which appears neither in $B$ nor in $D$ is not tagged. 
We switch learning off for those tokens, i.e., 
we mask the calculation of the loss and, thus, do not take them into account 
during optimization.
Raw sentences which do not contain at least one tagged word are discarded.

\textbf{Ambiguous annotation (\textbf{AMB})}
We next propose to annotate our raw sentences with \emph{noisy} or \emph{ambiguous} labels. We 
tag each token with all POS tags we consider possible, given dictionary $B$ and high-resource language data $D$, i.e.,  we assign to a token $w_l$ all tags in $\text{POS}(w_l)$ with  
\begin{align}
\text{POS}(w_l) = ~ \cup_{w_h\in \text{Tr}(w_l)} \text{POS}(w_h),
\end{align}
and all variables denoting the same quantities as before. 
In order to include our monolingual dictionaries, which give a set of 
possible tags $\text{POS}_{M}(w_l)$ for each word $w_l$ in the low-resource language, we extend this to
\begin{align}
\text{POS}(w_l) = ~ \cup_{w_h\in \text{Tr}(w_l)} \text{POS}(w_h) \cup \text{POS}_{M}(w_l)
\end{align}
Words that do not appear in either $M$, $B$ or $D$
are tagged with all possible tags. As before, we discard raw sentences which contain only tokens without information.

For training, we adapt the calculation of the cross entropy to ambiguous annotation. 
In particular, we obtain the loss of each example by treating the tag in $\text{POS}(w_l)$ which obtains the highest probability under our model as the gold label.

\section{Languages and Resources}
The goal of this work is to 
evaluate state-of-the-art unsupervised or weakly supervised POS-tagging strategies on truly low-resource languages.
For this, we select 15 languages from the Universal Dependencies (UD) project v2.1 \cite{ud2}.\footnote{\url{http://universaldependencies.org/}} All chosen languages are low-resource languages: 10 of them (Belarusian (be), Buryat (bxr), Upper Sorbian (hsb), Armenian (hy), Kazakh (kk), Kurmanji (kmr), Lithuanian (lt), Marathi (mr), Tamil (ta) and Telugu (te)) have less than 10k tokens for training in UD, and the other 5 languages (Amharic (am), Breton (br), Faroese (fo), Maltese (mt) and Tagalog (tl)) have no training data at all. Note, however, that we consider the training sets only to determine languages which are low-resource languages, and we do not make use of {\em any} training data in our experiments.
Our languages represent 6 different languages families: Afro-Asiatic (am and mt), Austronesian (tl), Dravidian (ta and te), Indo-European (be, br, fo, hsb, hy, kmr, lt, and mr), Mongolic (bxr), and Turkic (kk); cf. 
Table~\ref{tab-all-stats}. 

English is used as the high-resource language for cross-lingual transfer in our experiments. Note that a more informed choice of source language might be able to obtain better results, but that bilingual dictionaries are more likely to exist from/to English. 

\subsection{Dictionaries} \label{subsec-dic}

\textbf{Bilingual dictionaries}
A possible resource for weakly supervised approaches to cross-lingual POS tagging are bilingual dictionaries that contain word-to-word translations. They can be used for transferring information from resource-rich languages to low-resource languages, either by replacing words directly, transferring statistics, or for inducing cross-lingual word representations as done, e.g., by \citet{faruqui2014vec}. 
Some bilingual dictionaries can be downloaded from the \textsc{Wiktionary} user page of Matthias Buchmeier\footnote{\url{https://en.wiktionary.org/wiki/User:Matthias_Buchmeier}} or from the \textsc{Wikt2Dict} project website\footnote{\url{https://github.com/juditacs/wikt2dict}}. 
Unfortunately, those dictionaries 
do not include many low-resource languages (only 1 (lt) in our 15 languages is covered).
We therefore decide to rely on freely available resources for extracting bilingual dictionaries:  
\textsc{Panlex Swadesh}\footnote{\url{https://panlex.org}} and \textsc{Wikidata}\footnote{\url{https://www.wikidata.org}}.
The \textsc{Panlex Swadesh} Corpora \cite{baldwin2010panlex} gathers lists of 317 words (and synonyms) for over 600 languages. From this, we retrieve small bilingual dictionaries 
for almost all low-resource languages we consider.\footnote{\textsc{Panlex Swadesh} data is not available for kmr and mr.}
Our second resource, \textsc{Wikidata}, is an online database collecting more than 28 million links to \textsc{Wikipedia} and \textsc{Wiktionary} pages across different languages. The English pages align the multilingual sites, which enables us to extract bilingual dictionary entries. 
We extract bilingual dictionaries (English to target language) from these pages for our 15 low-resource languages.\footnote{We use the Mongolian and the Kurdish version for, respectively, Buryat and Kurmanji, which are closely related languages.} 
Statistics are provided in Table~\ref{tab-all-stats}. 
\\
\textbf{Monolingual dictionaries} 
A common approach for obtaining monolingual tag dictionaries, i.e., lists of tokens matched with their possible POS tags, for low-resource languages is to extract this information from \textsc{Wiktionary},\footnote{\url{https://www.wiktionary.org}} a multilingual free online dictionary. 
However, as an online collaborative tool that can be edited by any user, it is noisy and prone to errors. This is particularly true for low-resource languages, which is one of the reasons why studies such as 
\citet{li2012wiki} and \citet{wisniewski2014cross} limited themselves to a few (not truly resource-poor) languages.

The \textsc{Unimorph} project \cite{Kirov16wikitonary}, in contrast, was professionally created and contains morpho-syntactic information for 107 languages. 
However, this resource is only available for 3 of our languages (see Table \ref{tab-all-stats}), and POS tags are restricted to nouns, verbs, adjectives, as well as a small number of adverbs. 
We convert the \textsc{Unimorph} tagging scheme 
into UD POS tags for all languages.

\subsection{Embeddings}
We leverage monolingual word embeddings to improve our ability to generalize to unseen words in the target language. This can be seen as MTL with a language modeling auxiliary task. However, since the use of word embeddings is de facto standard in NLP, we consider them a basic part of our model. We employ the \textsc{Polyglot} embeddings which have been built from \textsc{Wikipedia} texts and made available by \citet{AlRfou13polyglot}.\footnote{\url{https://sites.google.com/site/rmyeid/projects/polyglot}} 
See Table~\ref{tab-all-stats} (last column) for the number of words covered by those embeddings. 

\section{Experiments}

We explore a 
set of state-of-the-art unsupervised and weakly supervised POS tagging strategies.
Note that we do not experiment with projected information from parallel corpora like  \citet{agic2015bible}, due to datasets only being available for 7 out of our 15 languages. As mentioned in the disclaimer above, we also do not consider limited supervision or contextualized word embeddings. 
\\
\textbf{\textbf{CHR11}} 
We employ the fully unsupervised method of \citet{Christodoulopoulos2011Bayesian}\footnote{\url{https://github.com/christos-c/bmmm}} (cf. related work section) 
as our first baseline.
We learn clusters on raw text extracted from \textsc{Wikipedia} using their Bayesian multinomial mixture model. 
To infer POS tags, we use our monolingual dictionaries to match each cluster with the POS tag that is most frequently associated with its tokens.
\\
\textbf{\textbf{GAR13}} Second, we compare to \citet{garrette2013learning}'s system\footnote{\url{https://github.com/dhgarrette/low-resource-pos-tagging-2014}} (cf. related work section), 
which consists of a maximum entropy Markov model which learns from a corpus of noisily labeled sentences. 
For this, we make use of the raw \textsc{Wikipedia} texts and our monolingual dictionaries.
\\
\textbf{\textbf{PLA16}}
\citet{Plank16multi} is a hierarchical LSTM POS tagger relying on a combination of character embeddings and word embeddings. \citet{Plank16multi} train this architecture in a multi-task fashion, using the task of predicting the log frequency of the next word as an auxiliary task. 
We compute log frequencies for each word $w_n$ as $int(\log(freq_{train}(w_{n+1})))$. 
We then learn POS taggers in a MTL setup: the main task learns to predict POS tags using the {ambiguous learning} strategy (data is created the same way as for AMB), and the auxiliary task learns a simplified language model.
The auxiliary task training data consists of sentences from \textsc{Wikipedia}.\\ 

\noindent We mostly adopt the hyperparameters from \citet{Plank16multi}.
The number of dimensions of our word embeddings is  64 (which is also the dimension of the \textsc{Polyglot} embeddings). 
We use one hidden layer of dimension $100$ for both the word and the character LSTM. In our use of dropout we also follow \citet{Plank16multi}. However, we further add character dropout 
with a coefficient of $0.25$ to improve regularization.
We train a minimum and maximum number of $15$ and $30$ epochs, respectively, and terminate if no improvement in the training loss is detected for $3$ consecutive epochs. At test time, we use the model 
which obtained the lowest loss.\\

\noindent We compute global accuracy scores over POS tags for all systems. 
Our evaluation is token-based, and each reported result is an average over $5$ runs
with different random seeds.

\section{Results}

The results of all experiments are presented in Table~\ref{tab-results-unsup};
the first three columns show the baseline results, 
the last four columns contain our proposed approaches. 
Our most important observation is that none of the approaches perform well on these languages. In fact, on average, no method gets even half of the tags right. This shows there is still a lot of room for exploring new and radically different approaches to learning POS taggers for (truly) low-resource languages. In addition, we make the following four observations: 
{\bf (i)} We see that the unsupervised baseline, CHR11, obtains the lowest accuracy on average over all languages. This shows that, while all POS taggers are poor, distant supervision from tag dictionaries may provide some signal. {\bf (ii)} On average over all languages, both single-task cross-lingual neural approaches, i.e., AMB and FREQ, outperform all baselines. Further, the difference between AMB and FREQ is with $0.0032$ small. Thus, it seems that 
both cross-lingual strategies, i.e., ambiguous labeling and using the most frequent tag as the gold label, work similarly well on average. Looking at individual languages, 
however, up to around $0.1$ performance difference can be found for am, hsb, hy, kmr, and te. Most likely, this 
difference might be explained by the frequency difference between the most frequent and other possible tags in the respective languages. {\bf (iii)} Looking at PLA16, we find that a language modeling auxiliary task seems to hurt performance on average. PLA16 uses the ambiguous annotation strategy for cross-lingual transfer and should, thus, directly improve over AMB. However, it only performs better for three languages: am, ta, and tl. This contrast to  \citet{Plank16multi}'s  results could be due to the small corpus sizes and, thus, mistakes in estimating  word frequencies. 
{\bf (iv)} The approaches with character-level MTL, i.e., AMB+AE and FREQ+AE, both improve over the respective single-task approaches. AMB+AE also obtains the highest average accuracy overall. Thus, in contrast to the language modeling auxiliary task, a character-level multi-task approach improves model performance in our setting.

\begin{table*}[h]
\centering
{\small
{\setlength{\tabcolsep}{9pt}
\begin{tabular}{@{~~}l|ccc@{~~~}c|c@{~~~}cc@{~~~}c|c@{~~~}cc@{~~~}c@{~~}}
\toprule
\textbf{Language} & \textbf{CHR11} & \textbf{GAR13} & \multicolumn{2}{c|}{\textbf{PLA16}} & \multicolumn{2}{c}{\textbf{AMB}} & \multicolumn{2}{c|}{\textbf{FREQ}} & \multicolumn{2}{c}{\textbf{AMB+AE}} & \multicolumn{2}{c}{\textbf{FREQ+AE}} \\
\midrule
am & \textbf{0.3441} & 0.1392 & 0.1643 & (0.00) & 0.1595 & (0.03) & 0.2479 & (0.00) & 0.1651 & (0.00) & 0.2608 & (0.01)\\
be & 0.2366 & 0.3524 & 0.4234 & (0.03) & 0.4627 & (0.03) & 0.4805 & (0.01) & 0.4285 & (0.02) & \textbf{0.5027} & (0.02)\\
br & 0.3442 & 0.3119 & 0.3267 & (0.01) & \textbf{0.3449} & (0.02) & 0.3247 & (0.01) & 0.3375 & (0.02) & 0.3325 & (0.01)\\
bxr & 0.4432 & \textbf{0.5295} & 0.3140 & (0.09) & 0.3783 & (0.07) & 0.4114 & (0.02) & 0.4153 & (0.09) & 0.4372 & (0.01)\\
fo & 0.4048 & 0.5671 & 0.5928 & (0.01) & 0.5992 & (0.00) & 0.5559 & (0.02) & \textbf{0.6016} & (0.00) & 0.5341 & (0.01)\\
hsb & 0.1886 & 0.3657 & 0.3573 & (0.06) & \textbf{0.4306} & (0.01) & 0.3540 & (0.01) & 0.4125 & (0.03) & 0.3446 & (0.01)\\
hy & 0.3706 & 0.3821 & 0.4940 & (0.02) & \textbf{0.5131} & (0.01) & 0.4302 & (0.02) & 0.5061 & (0.01) & 0.4482 & (0.01)\\
kk & 0.451 & 0.4271 & 0.4801 & (0.07) & 0.4809 & (0.06) & 0.4524 & (0.02) & \textbf{0.5370} & (0.02) & 0.4469 & (0.02)\\
kmr & 0.3201 & 0.3501 & 0.3165 & (0.06) & \textbf{0.3898} & (0.01) & 0.3020 & (0.01) & 0.3865 & (0.00) & 0.2948 & (0.01)\\
lt & 0.383 & 0.4460 & 0.5251 & (0.01) & \textbf{0.5266} & (0.02) & 0.4813 & (0.01) & 0.5226 & (0.01) & 0.4857 & (0.02)\\
mr & 0.2522 & \textbf{0.3862} & 0.3670 & (0.00) & 0.3710 & (0.01) & 0.3781 & (0.01) & 0.3799 & (0.01) & 0.3808 & (0.01)\\
mt & 0.3126 & 0.3002 & 0.2208 & (0.03) & 0.2666 & (0.04) & 0.3326 & (0.01) & 0.2924 & (0.06) & \textbf{0.3544} & (0.02)\\
ta & 0.3275 & 0.2758 & 0.3302 & (0.04) & 0.3163 & (0.05) & 0.3193 & (0.01) & \textbf{0.3562} & (0.00) & 0.3259 & (0.01)\\
te & 0.5035 & 0.5062 & 0.4430 & (0.06) & 0.4746 & (0.01) & \textbf{0.5734} & (0.02) & 0.4888 & (0.01) & 0.5615 & (0.01)\\
tl & 0.2774 & \textbf{0.5274} & 0.4931 & (0.01) & 0.4924 & (0.01) & 0.5157 & (0.02) & 0.5075 & (0.04) & 0.4651 & (0.04)\\
\midrule
Average & 0.3440 & 0.3911 & 0.3899 & & 0.4138 & &  0.4106 & & \textbf{0.4225} & & 0.4117 & \\
\bottomrule
\end{tabular}
}
}
\caption{POS tagging accuracy. 
For all neural models, the standard deviation is given in parentheses.}\label{tab-results-unsup}
\end{table*}

\section{Error Analysis}
The most frequent POS tags in the test treebanks are NOUN ($24.10\%$ on average), PUNCT ($15.46\%$ on average), and VERB ($13.57\%$ on average). 
We suspect this to be the main reason why the cluster-based method CHR11 achieves competitive results:
it tends to tag a huge part of the tokens with the most frequent POS tags, i.e., NOUN, PUNCT, and VERB. However, it fails on less frequent ones, losing against the other approaches. 
Bi-LSTMs tag more tokens belonging to less frequent categories correctly.
However, a possible source of errors might be that the training sets we created for AMB and FREQ are fairly unbalanced:
while the percentage of NOUN is  $24.10\%$ in the treebank test sets, only $15.19\%$ of the tokens are assigned this tag for FREQ.
The VERB tag, in contrast, which makes, on average, up for $13.57\%$ of the tokens in the treebank test sets, is assigned to  $24.65\%$ of the 
words for FREQ.
Similarly, the PUNCT tag is much more frequent for FREQ: $37.41\%$, as compared to $15.46\%$ in the 
test sets. 
For AMB, a total of $70.93\%$ of the tokens have been tagged as possible NOUN. The next most frequent tags are VERB and PROPN with $69.87\%$ and, respectively, $61.34\%$. PUNCT is slightly less frequent than in the test sets: only $14.21\%$ of the words have been tagged as such.\footnote{Annotations for AMB are ambiguous, i.e., each token can have multiple tags, such that the percentages do not sum up to $1$.}

\section{Comparison with High-Resource Languages}
Finally, the result that state-of-the-art taggers do not perform particularly well for the languages in our experiments is more valuable if we can find a plausible explanation for that.
We, thus, repeat a subset of the main experiments for 5 high-resource languages: German~(de), Spanish~(es), Italian~(it), Portuguese~(pt) and Swedish~(sv). We compare two different settings: First, we employ comparable resources to the previous experiment~(-). Second, we use the dictionaries (named Wiki-ly) provided by \citet{li2012wiki}, which are extracted from \textsc{Wiktionary} and contain more reliable POS tags,\footnote{Note that \citet{li2012wiki} use the Universal POS tags of \citet{petrov2012universal}, which do not exactly correspond to UD POS tags. We map, respectively, the CONJ and ``." tags to CCONJ and PUNCT. No PROPN, AUX and SCONJ tags occur in the dictionaries.} and replace \textsc{Wikipedia} texts with UD texts for training~(+). We keep all hyperparameters the same as in the previous experiments. Again, we perform 5 training runs for all neural models and report average scores.
We evaluate the scores on the PUD test sets \cite{conllsharedtask2017} of the UD treebanks for which there is a Wiki-ly dictionary available. 

\begin{table}[t]
\centering
{\small
{\setlength{\tabcolsep}{8pt}
\begin{tabular}{l|cc|cc}
\toprule
 \textbf{Language} & \textbf{CHR11} & \textbf{GAR13} & \textbf{AMB} & \textbf{FREQ} \\
\midrule
de - & 0.29 & 0.42 & 0.47 & 0.44 \\
es - & 0.47 & 0.36 & 0.46 & 0.56 \\
it - & 0.39 & 0.32 & 0.41 & 0.48 \\
pt - & 0.49 & 0.34 & 0.37 & 0.53 \\
sv - & 0.32 & 0.42 & 0.43 & 0.49 \\
\midrule
Average - & 0.39 & 0.37 & 0.43 & 0.50 \\
\midrule
de + & 0.59 & 0.59 & 0.66 & 0.72 \\
es + & 0.63& 0.74 & 0.72 & 0.80 \\
it + & 0.67 & 0.75 & 0.74 & 0.75 \\
pt + & 0.62& 0.66 & 0.73 & 0.77 \\
sv + & 0.59 & 0.69 & 0.70 & 0.80 \\
\midrule
Average + & 0.62 & 0.69 & 0.72 & 0.77 \\
\bottomrule
\end{tabular}
}
}
\caption{High-resource POS tagging accuracy. ``-'': comparable data to previous experiments; ``+'': higher quality data.}\label{tab-results-high}
\end{table}

\textbf{Results and discussion} 
Results 
are shown in Table~\ref{tab-results-high}. Comparing the average results of the neural models with those for the low-resource languages, we find an improvement of $0.02$ for AMB and $0.09$ for FREQ, respectively. Thus, those approaches work slightly better for our high-resource languages, maybe because those are closer to English, our source language for cross-lingual transfer. However, when higher quality resources are used, tagging performance is even $0.31$ higher for AMB. Similarly, CHR11 and GAR13 improve strongly over their results for the resource-poor languages.
For FREQ, performance nearly doubles: it increases by $0.36$.
This highlights the importance of using high quality resources.

\section{Related Work}
POS tagging and other NLP sequence labeling tasks have been successfully approached using bidirectional LSTMs \cite{wang2015unified,Plank16multi}. 
Early work using such architectures relied on large annotated datasets, but \citet{Plank16multi} showed that bi-LSTMs 
are not as reliant on data size as previously assumed. 
Their approach obtained state-of-the-art results for POS tagging in several languages, 
which is why we build upon it.
\citet{rei:2017:Long} showed how an additional language modeling objective could improve performance for POS tagging. 
Neural networks make MTL via parameter sharing easy; thus,
different task combinations have been investigated exhaustively
\cite{sogaard-goldberg:2016:P16-2,Augenstein2017KBC}. 
An analysis of task combinations was performed by \citet{Bingel2017}. \citet{ruder2017sluice} presented a more flexible architecture, which learns what to share between the main and auxiliary tasks. 
\citet{Augenstein2018NAACL} combined MTL with semi-supervised learning for strongly related tasks with different output spaces.
However, work on combining sequence labeling main tasks and seq2seq auxiliary tasks is harder to find.
\citet{dai2015semi} pretrained an LSTM as part of a sequence autoencoder on unlabeled data to obtain better performance on a sequence classification task. However, they reported poor results for joint training. We obtain different results: an autoencoding seq2seq task is beneficial for low-resource POS tagging.  
Cross-lingual approaches have been used for a large variety of
tasks, e.g., automatic speech recognition
\cite{huang2013cross}, entity
recognition \cite{MengqiuWang2014}, 
or parsing
\cite{
sogaard:2011:ACL-HLT20112,Naseem1,TACL892}. In the realm of seq2seq models, work existis on cross-lingual
 machine translation \cite{dong-EtAl:2015:ACL-IJCNLP2,
TACL1081}. Another example 
is a character-based approach by
\citet{kann2017one} for morphological generation.

\section{Conclusion}
We analyzed 
state-of-the-art approaches for low-resource POS tagging of {\em truly} low-resource languages: POS tagging in these languages is still difficult because resources are limited and of poor quality, with average tagging accuracies well below 50\%. Weakly supervised approaches only slightly outperform a state-of-the-art unsupervised baseline.

\section{Acknowledgments} 
We would like to thank Barbara Plank, Isabelle Augenstein, and Johannes Bjerva for conversations on this topic. AS is funded by a Google Focused Research Award.

\fontsize{9.0pt}{10.0pt} \selectfont
\bibliography{aaai}
\bibliographystyle{aaai}
\end{document}